\documentclass[letterpaper,10pt,journal,twoside]{IEEEtran}

\IEEEoverridecommandlockouts

\usepackage{cite}
\usepackage{amsmath,amssymb,amsfonts}
\usepackage{algorithmic}
\usepackage{graphicx}
\usepackage{textcomp}
\usepackage{xcolor}
\usepackage{booktabs}
\usepackage{subcaption}
\usepackage{wasysym}
\usepackage{multirow}
\usepackage{siunitx}
\usepackage{float}
\usepackage{pifont}
\usepackage[hidelinks]{hyperref}

\title{\LARGE \bf ORBIT: A Modular Multi-finger Robotic Hand Platform with Rotational Finger Reconfiguration
}
\author{
Benito Ribadeneira, Siqi Shang, Lillian Chin%
\thanks{This work has been submitted to the IEEE for possible publication. Copyright may be transferred without notice, after which this version may no longer be accessible.}%
\thanks{All authors are with The University of Texas at Austin, TX, USA. Correspondence: benitoalexrp@utexas.edu}}

\newcommand{\orbit}{\textsc{orbit}}

\begin{document}
\maketitle
\thispagestyle{empty}
\pagestyle{empty}

\begin{abstract}
Despite the popularity of multi-fingered robotic hands for dexterous manipulation, there is no systematic platform to  compare different hand morphologies against one another. We present \orbit{} (Orbital Reconfigurable Base-actuated Integrated Theta-motion Hand), a modular platform designed to enable principled exploration of multi-fingered hand morphology. The core contribution of \orbit{} is a rotating knuckle mechanism, where independent fingers travel around a cylindrical palm via a belt drive. This enables \orbit{} to perform quick reconfiguration ($<$ \qty{1}{\second}) of finger layout without hand disassembly. \orbit is a robust system, with finger average angular velocity of \qty{1.57}{\radian\per\second} and a root finger force resistance of \qty{50}{\newton}. Using a three fingers instantiation of \orbit, we demonstrate an initial pilot grasping study, comparing anthropomorphic, isometric, and Barrett-style finger layouts across eleven objects of varied size, shape, and texture. We further demonstrate how \orbit's non-anthropomorphic orbital DOF enables the achievement of tasks that would be difficult to perform with fixed-knuckle designs, such as cube face reorientation, lid threading, and precision pinching of sub-centimeter objects. 
These results establish \orbit{} as a viable platform for direct ``apples-to-apples'' comparison of multi-fingered hand morphologies for specific manipulation tasks. \href{https://www.youtube.com/watch?v=zpsbDud820E&list=PLGu-an_M0EYVOoQ1bKcNYhdq1l7eOi-oP}{\textcolor{blue}{Video demonstration}}

\end{abstract}

\section{Introduction}

To achieve the dexterity seen in human and animal manipulation~\cite{five_finger_review, the_claw_giant,century_robot_hand}, roboticists have long been inspired by the multi-finger hand designs seen in nature to accomplish complex tasks such as in-hand object reorientation~\cite{Siqi_cube_rotation,Aron_dollar_orbit, Other_cube_reorientation}, tool-use manipulation~\cite{valve_one_hand,leap}, and assembly~\cite{Assembly_multifinger_hand}. This has led to a proliferation of different multi-finger hand designs with varied morphologies. Early systems such as the Utah/MIT Dextrous Hand established the case for tendon-driven designs~\cite{jacobsen_design_1986,ruka_hand,okada_object-handling_1979,main}, while subsequent work explored linkage-driven hands
~\cite{hirzinger_mechatronics_2000,m.c.m.herathUnderactuatedLinkageFinger2018,bhirangi_all_2023,lovchik_robonaut_1999}.
Popular research platforms soon emerged, such as the BarrettHand~\cite{BarretHand}, Allegro Hand~\cite{AllegroHand_RIGHT_FORMAT}, and LEAP Hand~\cite{leap}. Although all of these platforms were nominally anthropomorphic by placing thumbs opposite the majority of the fingers, the thumb placement and size differ significantly. 
More recently, non-anthropomorphic designs have explored how departing from the human hand template can unlock new manipulation capabilities. This includes hands with supernumenary fingers~\cite{finger_extra, extra_thum}, hands that place their fingers isometrically around the palm~\cite{spherical_aron_dollar,Spherical_retractable_dollar,odhner_stable_2015}, and hands that introduce fingers with unconventional Degrees of Freedom (DOF)~\cite{Roller_tip_v2,Belt_paralel_girpper,weird_3fingers,roller,reconfigurable_palm,previous_orbit, si_deltahands_2024}.

Given this multitude of multi-fingered hand designs, it is somewhat unsurprising that there is no common consensus on what the most optimal hand design is for manipulation. Even relatively simple questions like ``how many fingers should a robot have?'' remain open questions~\cite{hand_review, Review_paper_hands,new_review_dexhand}, let alone more complex questions such as ``are the extra DOFs afforded by non-anthropomorphic mechanisms worth the additional mechanical complexity?''~\cite{Beyong_anthropomorphism,question_for_nonantro}.

While each design paper describes the advantages of its chosen morphology in isolation, direct comparisons between hands with different morphologies using identical hardware, control, and experimental conditions are rare. 
Previous research generally focuses on designing a single multi-finger hand to satisfy a specific goal, such as close emulation of a human hand~\cite{lovchik_robonaut_1999, noauthor_shadow_2023}, minimizing the number of actuators~\cite{BarretHand, Nail, odhner_stable_2015}, or minimizing cost~\cite{shaw_leap_2023, bulgarelli_low-cost_2016}. Multiple studies talk about different hand variables to explore when designing a robotic hand or a gripper ~\cite{Thumb_study,Critical_review}, while others uses simulation as a controlled environment to optimize a particular design parameter. ~\cite{noauthor_lang2morph_nodate,bad_nonantro_application,bai_learning_2025}. To the best of our knowledge, no modular robotic hand platform has been reported that enables direct, controlled comparison of robotic hand design parameters. This lack of systematic comparison makes it difficult to evaluate and select a hand morphology that is most appropriate for a given scenario.

\begin{figure} [t]
    \centering
    \includegraphics[width=\linewidth]{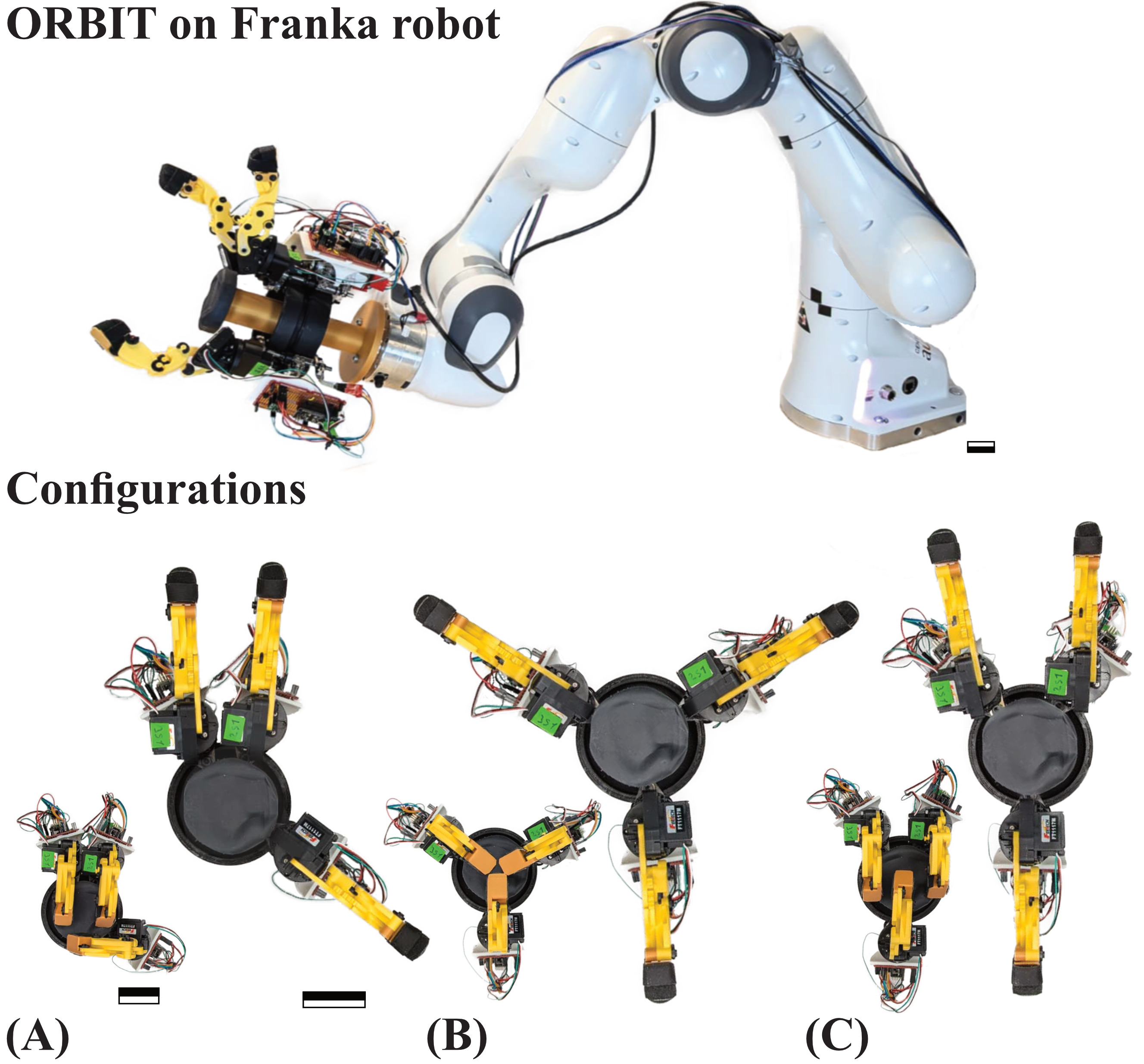}
    \caption {Overview of \orbit{} mounted on a robot arm. The same \orbit{} hardware allows us to use rotational movement to switch between different finger morphologies, including: (A) Anthropomorphic style, where two fingers oppose an offset thumb, (B) Isometric, where fingers are evenly distributed around the palm, (C) and Barrett style, where two fingers oppose a central thumb. Scale bars are \SI{50}{\milli\meter}.}
    \label{fig:Initial_figure}
    \vspace{-6mm}
\end{figure}

We address this research gap through \textbf{ORBIT}: a \textbf{O}rbital \textbf{R}econfigurable \textbf{B}ase-Actuated \textbf{I}ntegrated \textbf{T}heta-motion Hand. \orbit{} is a modular hand design in which multiple finger carriages independently rotate around a cylindrical palm via a belt-driven V-rail system (Fig.~\ref{fig:Initial_figure}). Each carriage serves as an independent mounting platform for a finger module, %
creating a modular architecture where nearly every relevant morphological variable can be changed without rebuilding the hand. The number of carriages determines the finger count, the finger module mounted on each carriage determines the finger geometry, and the rotational position of each carriage determines the finger layout (Fig.~\ref{fig:Initial_figure}-bottom). \orbit{} thus provides a viable platform for direct ``apples-to-apples'' comparison of multi-fingered hand morphologies.

In this paper, we discuss \orbit{}'s design and present a bio-inspired  instantiation with three 2-DOF's fingers and an anthropomorphic palm (Fig.~\ref{fig:Conceptual_pic}). We characterize the modular architecture of \orbit{} by demonstrating finger addition, removal and swapping. We also  characterize the rotating finger mechanism and current hand instantiation. \orbit{}'s fast finger displacement (\qty{1.57}{\radian\per\second}) and strong finger root force resistance (\SI{50}{\newton}) means that any finger layout can be achieved in less than one second and maintained under substantial load. Finally, we use \orbit{}'s angular finger displacement to demonstrate an initial investigation of three hand layout morphologies: anthropomorphic, isometric, and Barrett-style (Fig.~\ref{fig:Initial_figure}). We conduct qualitative grasp experiments with 11 objects and demonstrate three manipulation tasks that exploit \orbit{}'s non-anthropomorphic orbital degree of freedom: cube face reorientation, jar-lid unscrewing, and precision pinching of a sub-centimeter object. This paper makes the following contributions:

\begin{enumerate}
    \item A rotating knuckle mechanism that provides continuous, independently actuated root finger displacement around the palm, enabling rapid, in-situ reconfiguration of finger layout without hand disassembly.
    \item The design and characterization of a bio-inspired three finger hand design using \orbit{} demonstrating both precision and power grasping. 
    \item A controlled layout configuration comparison across three finger configurations and eleven objects, identifying task conditions under which each configuration performs best.
    \item Orbital manipulation demonstrations highlighting the contribution of dynamic finger-root displacement in challenging tasks, including cube reorientation, jar-lid threading, and sub-centimeter precision grasping.

\end{enumerate}

\section{Design}

\begin{figure} [t]
    \centering
    \includegraphics[width=\linewidth]{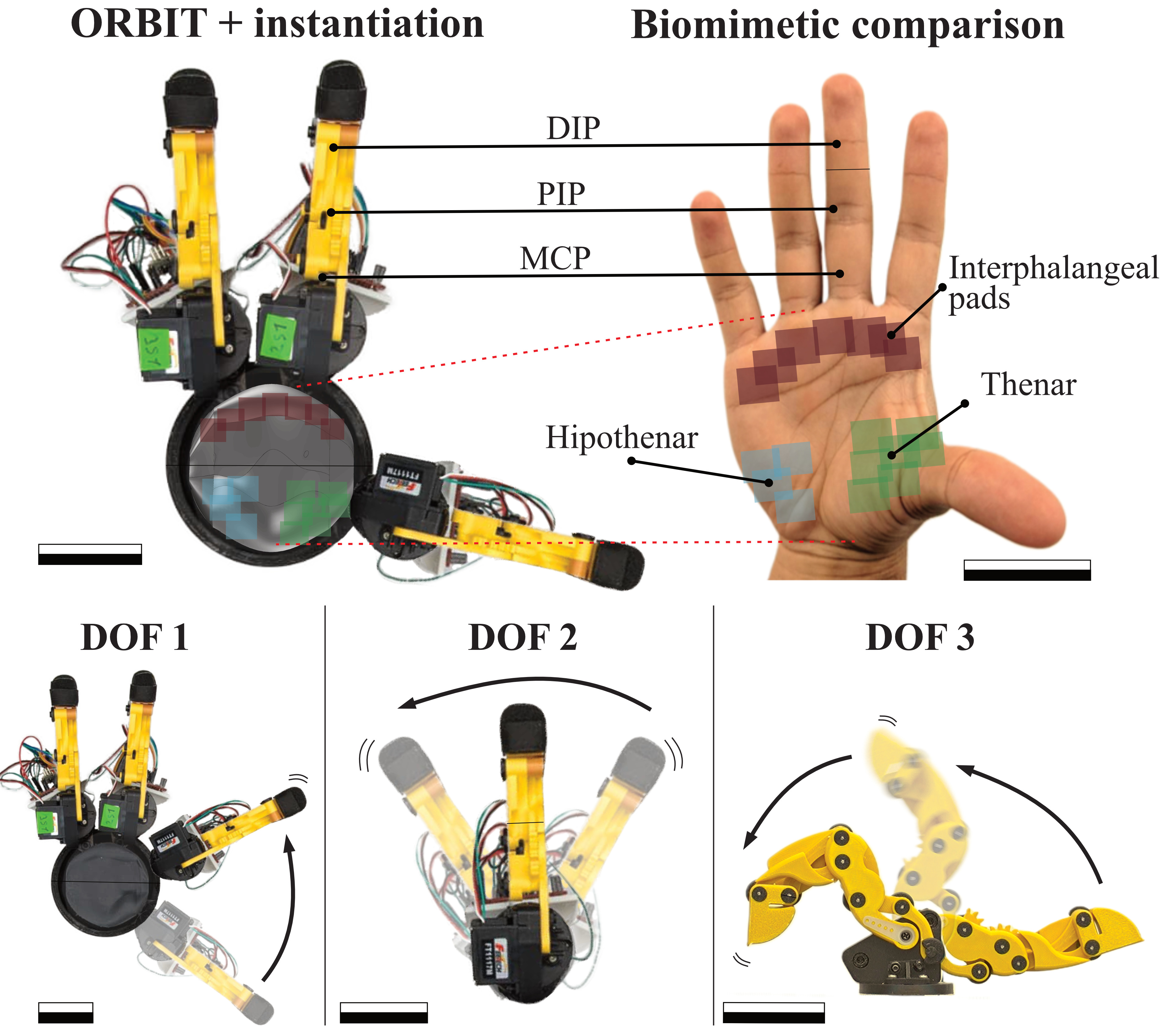}
    \caption{Conceptual overview of \orbit{} and its anthropomorphic hand morphology instantiation. Each of \orbit{}'s DOFs have a biological analogue: (1) root finger displacmeent, (2) abduction / adduction, and (3) flexion / extension. Similarly, the finger joints correspond to the Distal interPhalangeal (DIP), Proximal InterPhalangeal (PIP), MetaCarpoPhalangeal (MCP) joints of the finger, while  the soft palm mimics the thenar, hipothenar and Interphalangeal pads on the human palm. Scale bars are \SI{50}{\milli\meter}}
    
    \label{fig:Conceptual_pic}
    \vspace{-5mm}
\end{figure}

\begin{figure*}[!t]
  \centering
  \includegraphics[width=\linewidth]{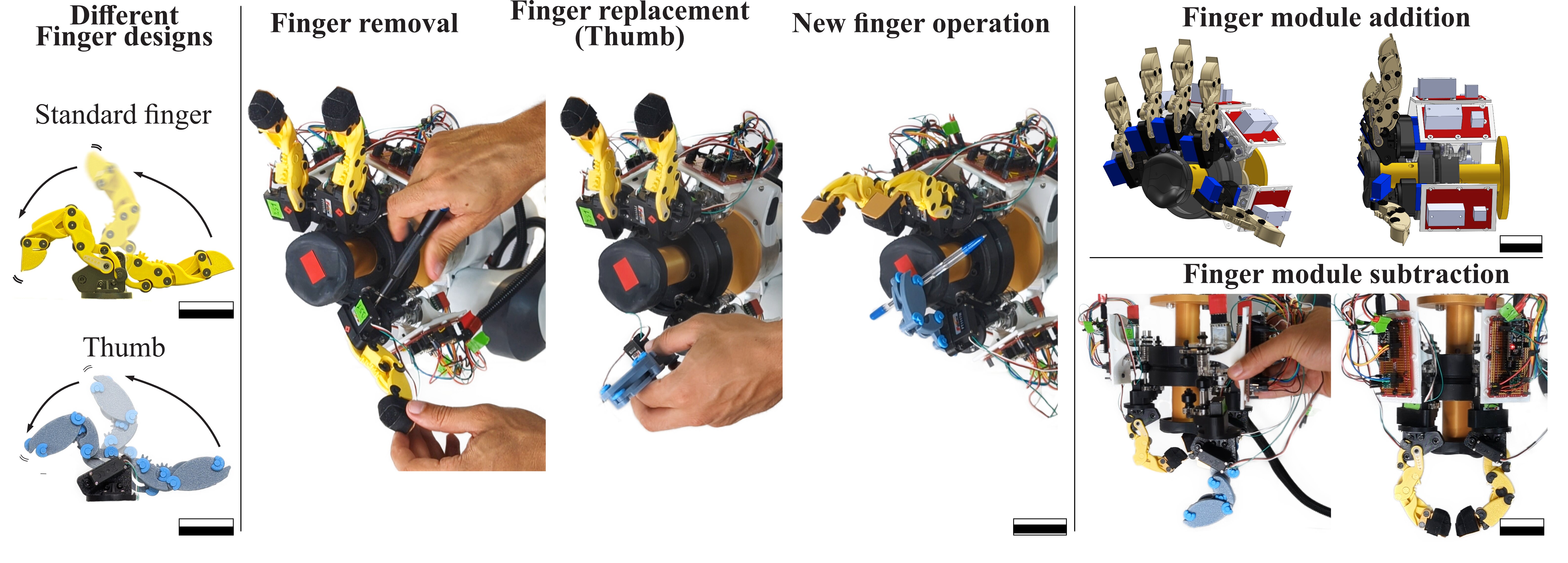}
  \caption{Demonstration of \orbit{}'s modular architecture. From left to right, we demonstrate how we can swap a standard finger design with a shorter thumb finger design without needing to disassemble the entire hand. We also demonstrate conceptual pictures of how additional finger modules can be added or removed from the rotating knuckle mechanism, creating new hand layouts with the same base architecture. Scale bars are \SI{50}{\milli\meter}}
  \label{fig: Modularity}
  \vspace{-5mm}
\end{figure*}

Our design goal for \orbit{} is to create a multi-finger platform that enables 
\begin{enumerate}
    \item Dynamic changing of finger layout without re-designing the entire hand.
    \item Modular architecture for finger addition, subtraction, and substitution
\end{enumerate}

To achieve these goals, we designed the core of \orbit{} to be a ``rotating knuckle'' mechanism, where independent finger carriages rotate around a central palm (Fig.~\ref{fig:Conceptual_pic}). Inspired by the non-anthropomorphic design recommendation in \cite{Beyong_anthropomorphism,nonantro_opinion}, we chose to not copy the static knuckles between the palm and the fingers seen in humans (metacarpophalangeal joints), but instead allowed the knuckles to rotate arbitrarily around the hand (Fig.~\ref{fig:Conceptual_pic}-DOF1), similar to~\cite{previous_orbit}. Unlike~\cite{previous_orbit}, we actuate each knuckle by a timing belt around a single cylindrical rail, which provides more compliance and flexibility than the prior dual track geared design. In this section, we discuss the design of the rotating knuckle mechanism and specific details of our chosen anthropomorphic hand instantiation (i.e. finger and palm).

\subsection{Rotating Knuckle Mechanism}
To create the rotating knuckle mechanism, we adapted the concept of a linear v-rail and carriage system to a cylindrical geometry. By having a platform serve as the static v-rail that the finger carriages rotate around, each finger can move independently and dynamically change the overall finger layout.
Traditional linear v-rail and carriage systems work by having a machined v-shaped groove serve as a track for a carriage. Each carriage has complementary v-shaped wheels that follow the groove. If we run a belt drive through the wheels, we can control the position of the carriage along the v-rail, similar to a carriage on a 3D printer gantry. We adapt this design to a palm / hand concept by making the v-rail circular rather than linear. Due to the lack of off-the-shelf cylindrical v-rail systems, we created a two part 3D printed cylindrical platform. We create the v-groove between two separate 3D printed parts (PLA (polylactic acid), Bambu Lab X1-Carbon), which screw onto each other to form a tight connection. By splitting the rail, we are able to print both parts without supports on the v-groove, enhancing smooth displacement of the carrier and minimizing tire wear.

\begin{figure*} [t]
    \centering
    \includegraphics[width=\linewidth]{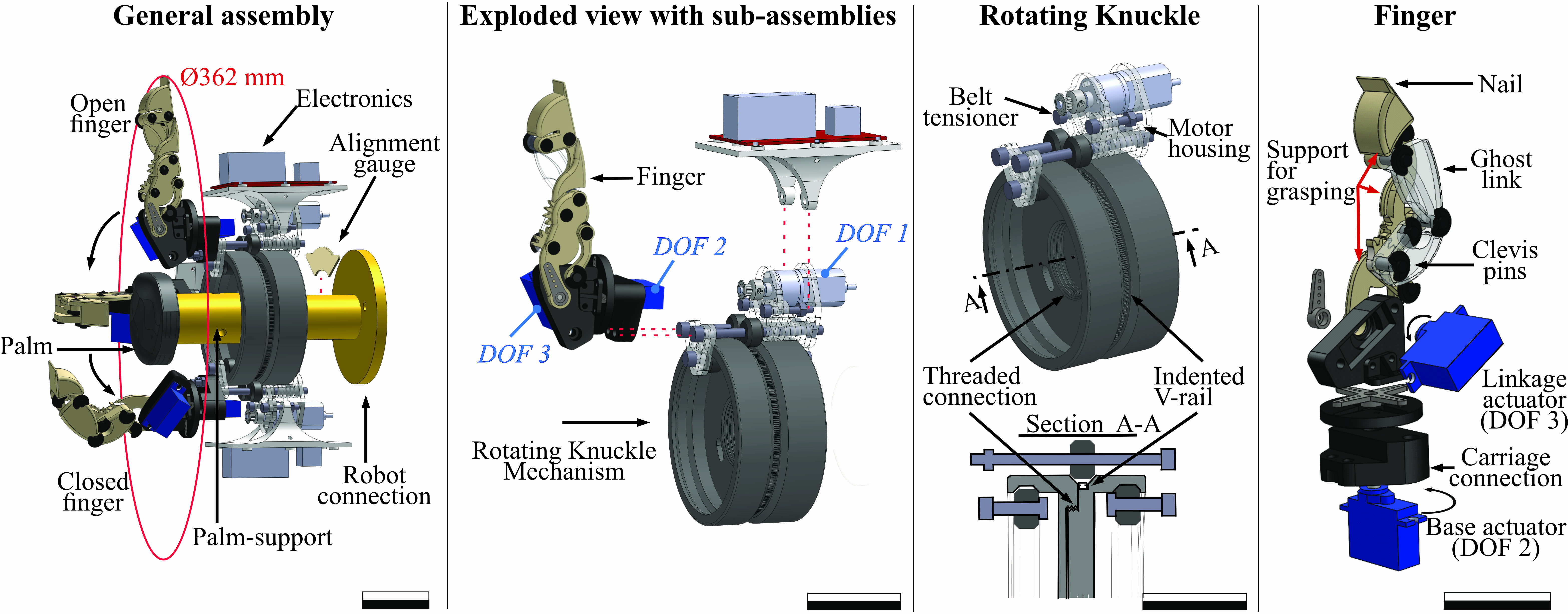}
    \caption{Overview of the specific \orbit{} instantiation we use in the paper. From left to right,  general overview of \orbit{} as a whole, exploded view of the two main subassemblies, and detailed information of the rotating knuckle mechanism and finger design. Scale bars are \SI{50}{\milli\meter}}
    \label{fig:Subassemblies}
    \vspace{-5mm}
\end{figure*}

To design the carriage, we extend the traditional v-rail carriage with additional wheels for added stability. Commercial linear V-rails usually have only one set of opposing wheels, as they are designed to work under unidirectional axial loading. However, in our use case of robot manipulation, the fingers must work under multidirectional force conditions and moments, requiring additional support. We thus add an additional two opposing v-wheels and rails in a triangular position to evenly absorb forces applied to the finger (Fig.~\ref{fig:Subassemblies}-Rotating Knuckle). We also indent the rail at the bottom of the external wheels according to the belt characteristics (MXL Timing Belt, 1/8" Width) to enable simultaneous movement of the fingers around the rail without slippage.

Each carriage is actuated by an onboard DC motor (Pololu 25:1 Metal Gearmotor) to drive the timing belt along the V-rail. The motor was chosen for its minimal footprint and ability to resist at least 10 \si{\newton} of force at stall for finger stability. All carriages use the same timing belt, with each carriage effectively serving as a belt tensioner for the other carriages. The belt's length was determined by CAD simulation of the minimum and maximum contour lengths according to the position of the motor pulley and the angle relationship of the motor shaft. 

\orbit{}'s modular design means that any finger design and layout can be chosen on top of the rotating knuckle mechanism, As demonstrated in Fig.~\ref{fig: Modularity}, the independent finger carriage design allows us to easily replace one finger module for another, as well as add / remove additional finger modules. To swap a finger, we only need to unplug electrical connections and unbolt the two-screw carriage interface. Meanwhile, adding / removing modules requires replacing the timing belt with a new appropriate length to maintain proper tension. Thus, achieving the modular architecture required for a morphology testing hand platform.

\subsection{Hand Instantiation}

    \begin{figure} [t]
    \centering
    \includegraphics[width=0.9\linewidth]{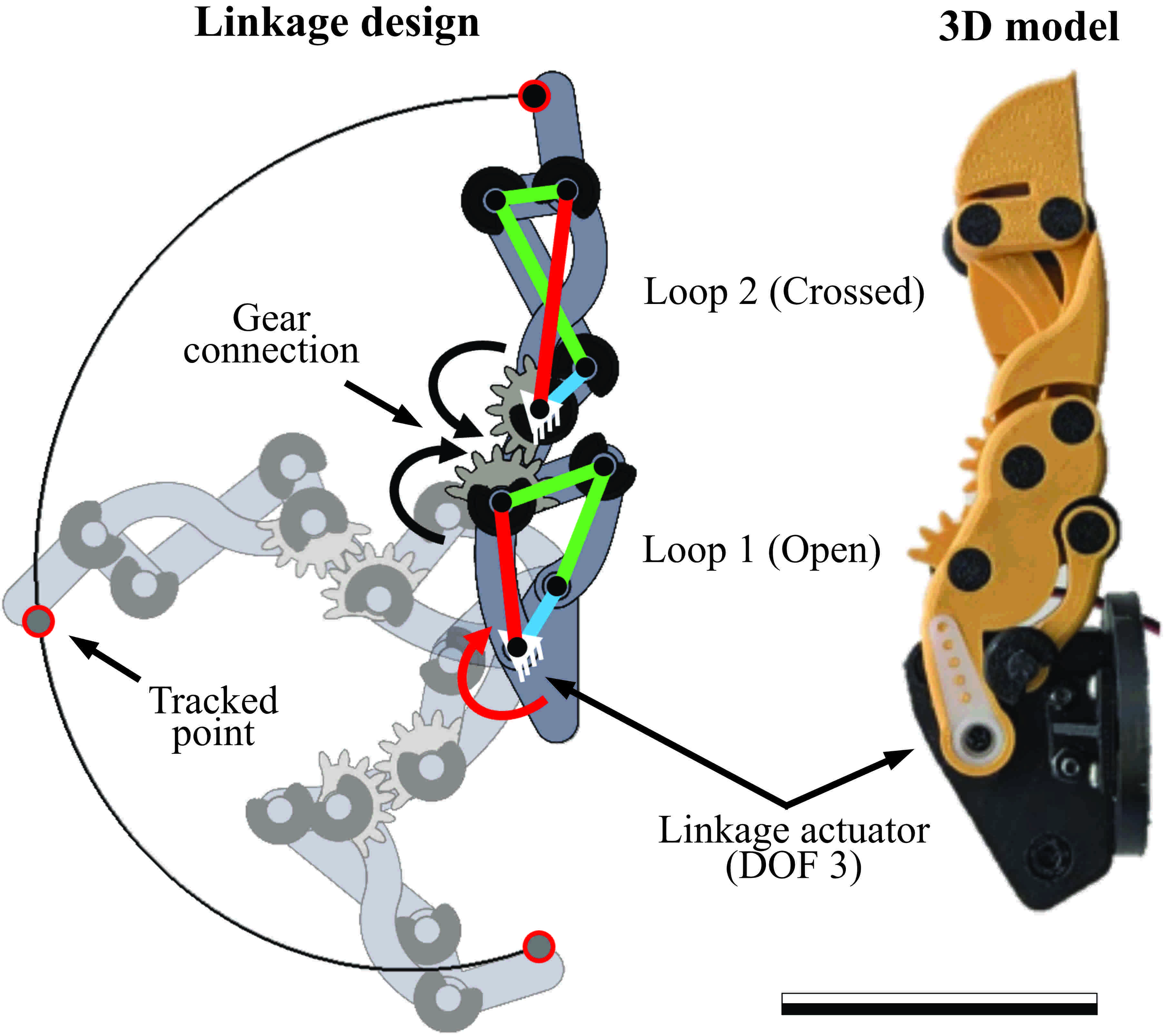}
    \caption{Kinematic diagram of the finger design with the two cascaded four-bar linkage. Red links are driven inputs, blue links are grounded, green links are couplers and $\theta$ is the servo-input angle. Scale bars are \SI{50}{\milli\meter}}

    \label{fig:finger exploited view}
    \vspace{-5mm}
    \end{figure}

To demonstrate \orbit{}'s functionality for this paper, we choose to explore the design space of a three-fingered hand uniformly comprised of 2-DOF anthropomorphic fingers. We chose three fingers as a reasonable trade off between mechanical simplicity and versatility \cite{question_for_nonantro}. Each finger body is a bio-inspired linkage system of two four bar linkages in series, connected by 3D printed clevis pins (Fig.~\ref{fig:finger exploited view}). This design is analogous to human morphology~\cite{linkage_finger_review}; one DOF mimics finger abduction/adduction through direct drive by a servo (FEETECH Mini Servo FT1117M-FB) between the finger and carriage (Fig.~\ref{fig:Conceptual_pic}-DOF2), while the other DOF uses a servo to drive linkages to mimic finger fexion/extension (Fig.~\ref{fig:Conceptual_pic}-DOF3). As detailed in Fig.~\ref{fig:finger exploited view}, the finger motion starts with the red link of the first loop powered by the servo and is transmitted to the second loop through the gear connection. Both loops can be solved using the standard four bar linkage equation~\cite{Design_of_machinery_book}. where the dimensions of all the links ($\ell$) are known and the input angle $\theta_1$ is given by the servo motor.

    At the fingertip, a rigid \SI{1}{\milli\meter} PLA nail extends past a \SI{5}{\milli\meter} compliant sponge pad covered with friction tape, enabling the finger to slide under and grip flat or sub-centimeter objects that a rounded pad alone would push away~\cite{Nail}. Overall, grasp compliance is achieved passively as the sum of three mechanical sources: slack in the servo joints, backlash in the spur gear coupling between the two linkage loops, and compression of the sponge pad at the fingertip, allowing all three fingers to conform to a common object geometry without force sensing.

    To support power grasping, we integrated a bio-inspired palm into the center of our hand (Fig.~\ref{fig:Conceptual_pic}). We replicated the soft prominences of the human hand (thenar, hypothenar, and palmar interphalangeal pads) in our artificial pad by 3D printing a soft shell using Formlabs Silicone 40A. We attach the soft palm to the rotating knuckles mechanism by surrounding a circular protrusion of a  rigid 3D printed PLA structural body.
    We connect the entire \orbit{} assembly to a robot arm (Franka Emika Panda) through a 3D printed PLA connection piece. Three alignment holes spaced at $120^\circ$ intervals on the connector body serve as docking points for a positioning gauge used to reset the carriage encoder to a known reference at startup. 
    
    To control \orbit{}, we follow the same modularity philosophy as the mechanical architecture. Each carriage is a self-contained unit carrying an Arduino Nano and a H-bridge DC motor driver on a breadboard, providing \SI{12}{\volt} to the DC motor and \SI{5}{\volt} to both servos. All three DOF are controlled by independent PID position loops running on the Arduino Nano. For DOF~1, the magnetic encoder on the DC motor shaft provides position feedback, while DOF~2 and DOF~3 are controlled by the servos' internal position loop with potentiometer feedback. Each Arduino communicates with the host computer via USB-serial, receiving target angle commands and confirming when the targets are reached.

\begin{table}[t]
  \centering
  \caption{Consolidated \orbit{} performance characterization.}
  \label{tab:char_summary}
  \renewcommand{\arraystretch}{1.2}
  \begin{tabular}{l      l      l}
    \toprule
    \textbf{DoF} & \textbf{Metric} & \textbf{Result} \\
    \midrule
    DoF~1 & Position accuracy & Within $\pm 2.0^\circ$ \\
    DoF~1 & Force resistance  & $49.46 \pm 0.9\,\si{\newton}$ \\
    DoF~1 & Angular velocity  & $1.57\,\si{\radian/\second}$ \\
    \midrule
    DoF~2 & Position accuracy & $< 2^\circ$ \\
    DoF~2 & Travel time       & $0.53 \pm 0.01\,\si{\second}$ \\
    \midrule
    DoF~3 & Position accuracy & $< 2^\circ$ \\
    DoF~3 & Travel time       & $0.96 \pm 0.09\,\si{\second}$ \\

    \bottomrule
  \end{tabular}
  \vspace{-5mm}
\end{table}

\section{Characterization}
We characterize \orbit{}'s rotating knuckle mechanism across three metrics: position accuracy, actuation speed, and force resistance. We also characterize the position accuracy and speed of the fingers. Results are summarized in Tab.~\ref{tab:char_summary}.

\begin{figure*} [t!]
    \centering
    \includegraphics[width=\linewidth]{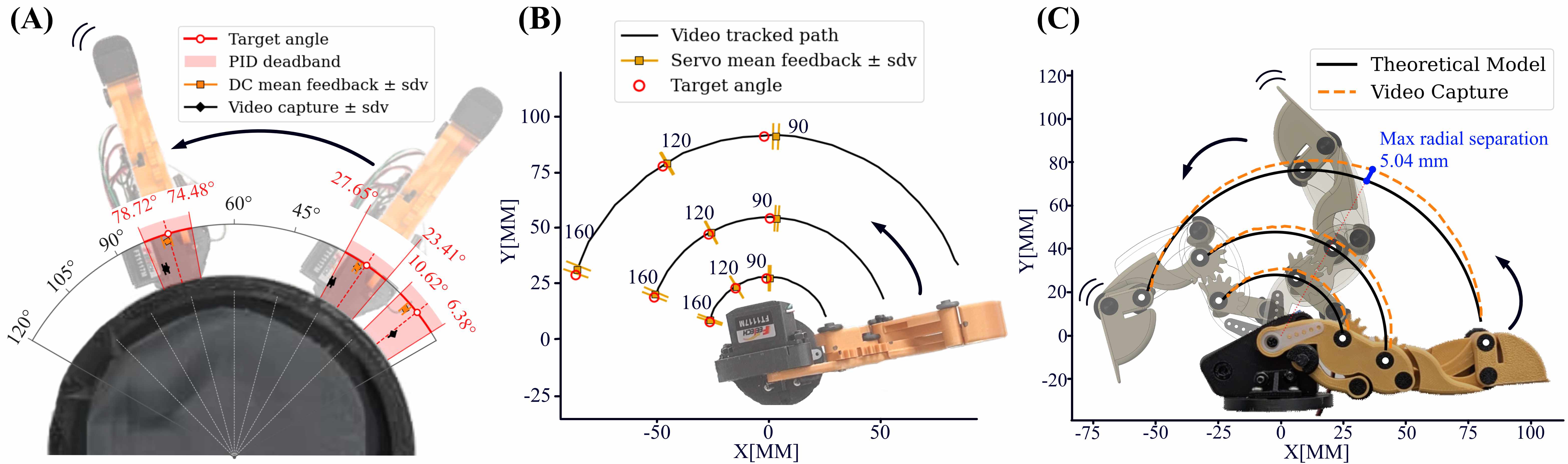}
    \caption{Position accuracy characterization of the three DOF's of \orbit{} (A) DOF~1 DC motor feedback vs video tracking vs target angle. (B) DOF~2 servo motor feedback vs target angle around the video tracked path. (C) DOF~3 Video tracking vs Theoretical model path.}
    \label{fig:General characterization}
    \vspace{-5mm}
\end{figure*}

\subsection{Rotating Knuckle Mechanism}

Each of the three carriages was characterized individually, with fingers mounted and the hand held in a fixed, gravity-neutral orientation on the robot arm. DOF~2 and DOF~3 of the finger were held at their midpoint angles to maintain a consistent inertial load across all trials.

\subsubsection{Position accuracy} Using the DC motor encoder, the motor shaft rotation is correlated with the carriage position around the palm. We sent step commands back and forth three times to target angles of $120^\circ$, $360^\circ$, and $1080^\circ$ of motor shaft ($8.5^\circ$, $25.5^\circ$, and $76.6^\circ$ of carriage angle), with a PID dead-band of $\pm 30^\circ$ motor shaft ($\approx \pm 2.0^\circ$ carriage angle). 
Fig.~\ref{fig:General characterization}A shows the encoder feedback and video tracking with its error bars against the step command. After achieving its steady-state (0.6 seconds), both signals reach the target within the PID dead-band on every transition, 

\subsubsection{Angular velocity} The travel time for the three configurations (anthropomorphic, isometric, Barrett-style) was measured from video, giving an average carriage angular velocity of $1.57\,\si{\radian/\second}$ (with a peak of $2.36\,\si{\radian/\second}$ at the largest displacement). A \qty{76.60}{\degree} transition, the largest displacement between any of the three evaluated finger layouts, completes in under \SI{0.6}{\second}, meaning \orbit{} can switch between any of the three configurations of Fig.~\ref{fig:Initial_figure} in less than one second.

\subsubsection{Force resistance.} A force perpendicular to the finger axis was applied and increased until the carriage displaced past the $30^\circ$ dead-band or the timing belt slipped (SI Video). The force gauge reached its \SI{50}{\newton} limit before any displacement was registered in nine out of ten trials, with the belt slipping only once at \SI{42}{\newton}. This confirms that the mechanism holds stable finger positions under expected grasping loads.

\subsection{Hand Instantiation}
\subsubsection{Position accuracy} To characterize the position accuracy, we mounted a single finger on a bench vise and evaluated motion across the full servo range ($20^\circ$--$160^\circ$). We compared the two DOF's servo encoder feedback against video tracking of three marker points along the finger. Since DOF~2 is directly driven, the encoder and all three markers demonstrated rigid coupling with no measurable flex (Fig.~\ref{fig:General characterization}B), leading to a maximum deviation below $2^\circ$. For DOF~3, the kinematic model computed from the servo angle was compared against the video capture (Fig.~\ref{fig:General characterization}C). The maximum fingertip error (\SI{5.04}{\milli\meter}) is seen at the extremes where joint backlash accumulates. 

\subsubsection{Actuation speed} We measure the travel time of both finger DOFs over the full $140^\circ$ servo range. Over 5 trials, DOF~2 travels in $0.53 \pm 0.01\,\si{\second}$ and DOF~3 in $0.96 \pm 0.09\,\si{\second}$. Although both DOFs use the same servo, DOF~3 is slower due to the added friction between the finger linkages and the deceleration required to reach the final fingertip position.

Overall, the kinematic model matches well against video tracking, enabling contact configurations to be planned without per-object calibration. We further note that root finger re-configuration completed in under 1 \si{\second}, allowing DOF~1 to be used as an active manipulation primitive. 

\section{Grasping performance and discussion}
    \label{sec:Grasping}

Using the three finger instantiation on \orbit{}, we evaluate its grasping performance by varying its finger layout configurations and performing grasping / manipulation tests with household objects. In our experiments, we focus on comparing three common hand configurations (Fig~\ref{fig:Initial_figure}:

\begin{enumerate}
    \item Anthropomorphic, where two side by side fingers oppose an offset thumb
    \item Isometric, where the fingers are evenly spaced around the palm
    \item Barrett-style, where two slightly spread fingers oppose a centered thumb
\end{enumerate}
To achieve each hand configuration, we set DOF~1 of each finger relative to the palm circumference to be at a fixed position; (\qty{20}{\degree}, \qty{60}{\degree}, \qty{270}{\degree}) for Anthropomorphic, (\qty{0}{\degree}, \qty{120}{\degree}, \qty{240}{\degree}) for Isometric, and (\qty{345}{\degree}, \qty{75}{\degree}, \qty{210}{\degree}) for Barrett, keeping DOF2 and DOF3 free for grasping. 

\subsection{Grasping Force}
\begin{figure}[t!]
  \centering
  \includegraphics[width=\linewidth]{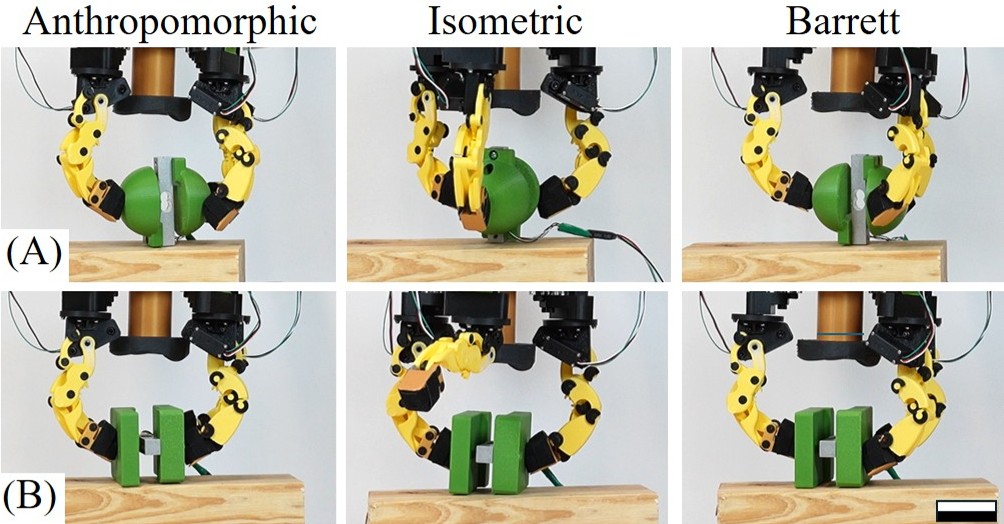}
  \caption{Grip force experimental setup for the three hand configurations with the two load cell presentations: (A) a spherical layout and (B) a flat layout. Scale bars are \SI{50}{\milli\meter}.}
  \label{fig:grip_force}
\end{figure}

To characterize each configuration's grip force, we commanded all three fingers  to maximum flexion (DOF~3 = \qty{20}{\degree}), with the abduction angle (DOF~2) pointing toward a load cell with different surface presentations (Fig.~\ref{fig:grip_force}). Specifically, we outfitted the load cell with a spherical surface (25 mm diameter) placed at the geometric center of the three fingertips, and a flat surface (20 mm × 40 mm), pressed by as many fingers as possible. For the isometric test, this meant only grasping with two opposing fingers, as we found that a three finger grasp led to two fingers grabbing the edges of the flat cell, resulting in falsely weaker results.  

Table~\ref{tab:grip_force} reports the results of the grip force after a \SI{10}{\second} settling window. The flat load cell reports higher forces, as all three finger forces are applied perpendicular to the gauge. Meanwhile, for the sphere, at least one finger ends up applying a non perpendicular force, causing smaller measurements of the overall grasp force. Across all three cases, the anthropomorphic configuration presents the highest grip force because of its root finger separation. The two adjacent opposing fingers transmit a more vertical force to the center of the gauge, increasing the readings as the separation between finger roots decreases.
 
\begin{table}[t!]
  \centering
  \caption{Grip force (N) for each hand configuration and load case after
  a \SI{10}{\second} settling window. Mean $\pm$ standard deviation over three trials}
  \label{tab:grip_force}
  \renewcommand{\arraystretch}{1.25}
  \begin{tabular}{l c c c}
    \toprule
    \textbf{Load case} & \textbf{Anthropomorphic} & \textbf{Isometric} & \textbf{Barrett} \\
    \midrule
    Spherical & $2.02 \pm 0.18$ & $1.64 \pm 0.15$ & $1.47 \pm 0.37$ \\
    Flat      & $2.42 \pm 0.16$ & $1.66 \pm 0.16$ & $2.20 \pm 0.21$ \\
    \bottomrule
  \end{tabular}
  \vspace{-5mm}
\end{table}

\subsection{Object Grasping Evaluation}
\begin{figure}[t!]
  \centering
  \includegraphics[width=\linewidth]{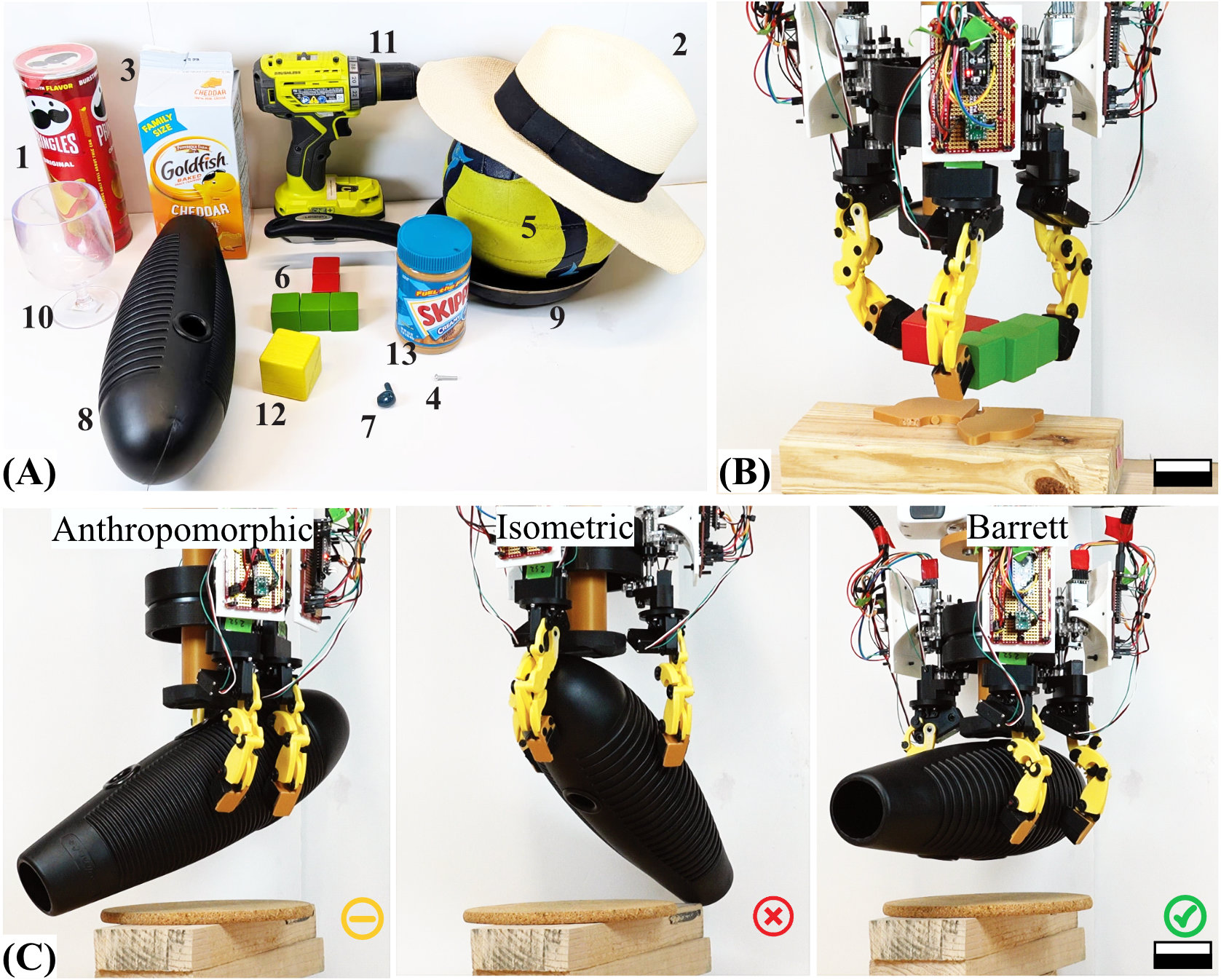}
  \caption{(A) Manipulation target overview. Objects 1-11 were used to test grasp success. In numerical order, the objects are: Pringles can, hat, cracker box, M3x\qty{5}{\milli\meter} screw, volleyball, five \qty{2}{\centi\meter} cubes, earbud, guiro, pan, glass, drill. Objects 12 (\qty{5}{\centi\meter} cube) and 13 (peanut butter jar) were used to demonstrate the non-anthropomorphic ability of finger displacement. (B) Demonstration of five cubes lifted under isometric layout.
  (C) Guiro pick-up with the three configurations, showing a partially accepted, a failed and a successful pick-up, respectively. Scale bars are \SI{50}{\milli\meter}}
  \label{fig:objects_grasping}
\end{figure}

We evaluated how well different hand configurations grasp by testing each hand layout against 11 objects (Fig.~\ref{fig:objects_grasping}A). For each object-layout pair, we lowered the arm to contact the object, closed DOF~3 to a position hard-coded per object, and lifted the object by \SI{100}{\milli\meter}. Trials were marked as a success if the object was held stably for \SI{3}{\second} (Tab.~\ref{tab:grasp_results}).

The anthropomorphic configuration performed best on objects with uneven mass distribution (i.e., pan and drill), where its offset thumb resists the overturning moment produced by the object's weight. The isometric configuration performed best on large symmetric objects (i.e., volleyball and the five small cubes), where its $120^\circ$-spaced contacts triangulate at the object center and produce a balanced grip force distribution. Finally, the Barrett configuration performed best for tool objects with centered mass (i.e., guiro), where its centered thumb enables a leveled pick-up when grasping objects by their center. The aligned opposite thumb of Barrett makes this configuration share some of the benefits of the other two configurations, as the Barrett hand could partially lift the volleyball (like the isometric) and lift the pan with a slightly centered grasp (like the anthropomorphic). This intermediate ability level is largely due to the wide abduction/adduction range of DOF~2, which lets the Barrett layout emulate the other configurations without finger displacement. 

All three configurations grasped everyday objects (Pringles bottle, hat, earbud, glass, drill) successfully, demonstrating hardware robustness. Challenging objects for all configurations included the \SI{430}{\gram} cracker box, which exceeded the fingertip friction and servo grasp force, and the M3x5mm screw, which required a fine pinch grasp that could not be performed when the finger roots were fixed.

\begin{table}[t!]
  \centering
  \small
  \caption{Grasping success for each hand configuration. Objects are described in Fig.~\ref{fig:objects_grasping}. A = Anthropomorphic, I = Isometric, B = Barrett. \ding{51} = success, \ding{55} = failure.}
  \label{tab:grasp_results}
  \renewcommand{\arraystretch}{1.2}
  \begin{tabular}{c l c c c c c}
    \toprule
    \textbf{Object} & \textbf{Challenge} &
    \textbf{A} & \textbf{I} & \textbf{B} \\
    \midrule
    1  & Tall, light     & \ding{51} & \ding{51} & \ding{51} \\
    2  & Large brim      & \ding{51} & \ding{51} & \ding{51} \\
    7  & Small, smooth   & \ding{51} & \ding{51} & \ding{51} \\
    10 & Delicate        & \ding{51} & \ding{51} & \ding{51} \\
    11 & Heavy           & \ding{51} & \ding{51} & \ding{51} \\
    6  & Unstable        & Four & \ding{51} & \ding{51} \\
    8  & Large, smooth   & Not level & \ding{55} & \ding{51} \\
    9  & Uneven mass     & \ding{51} & \ding{55} & Off-center \\
    5  & Even mass       & \ding{55} & \ding{51} & Partial \\
    3  & Flat faces      & \ding{55} & \ding{55} & \ding{55} \\
    4  & Sub-cm          & \ding{55} & \ding{55} & \ding{55} \\
    \midrule
    \multicolumn{2}{l}{\textbf{Success (including partial)}} & 
    \textbf{8} & \textbf{7} & \textbf{9} \\
    \bottomrule
  \end{tabular}
  \vspace{-5mm}
\end{table}

\subsection{Qualitative Manipulation Demonstrations}
 \label{subsec:Demonstrations}
 
\begin{figure}[t!]
  \centering
  \includegraphics[width=\linewidth]{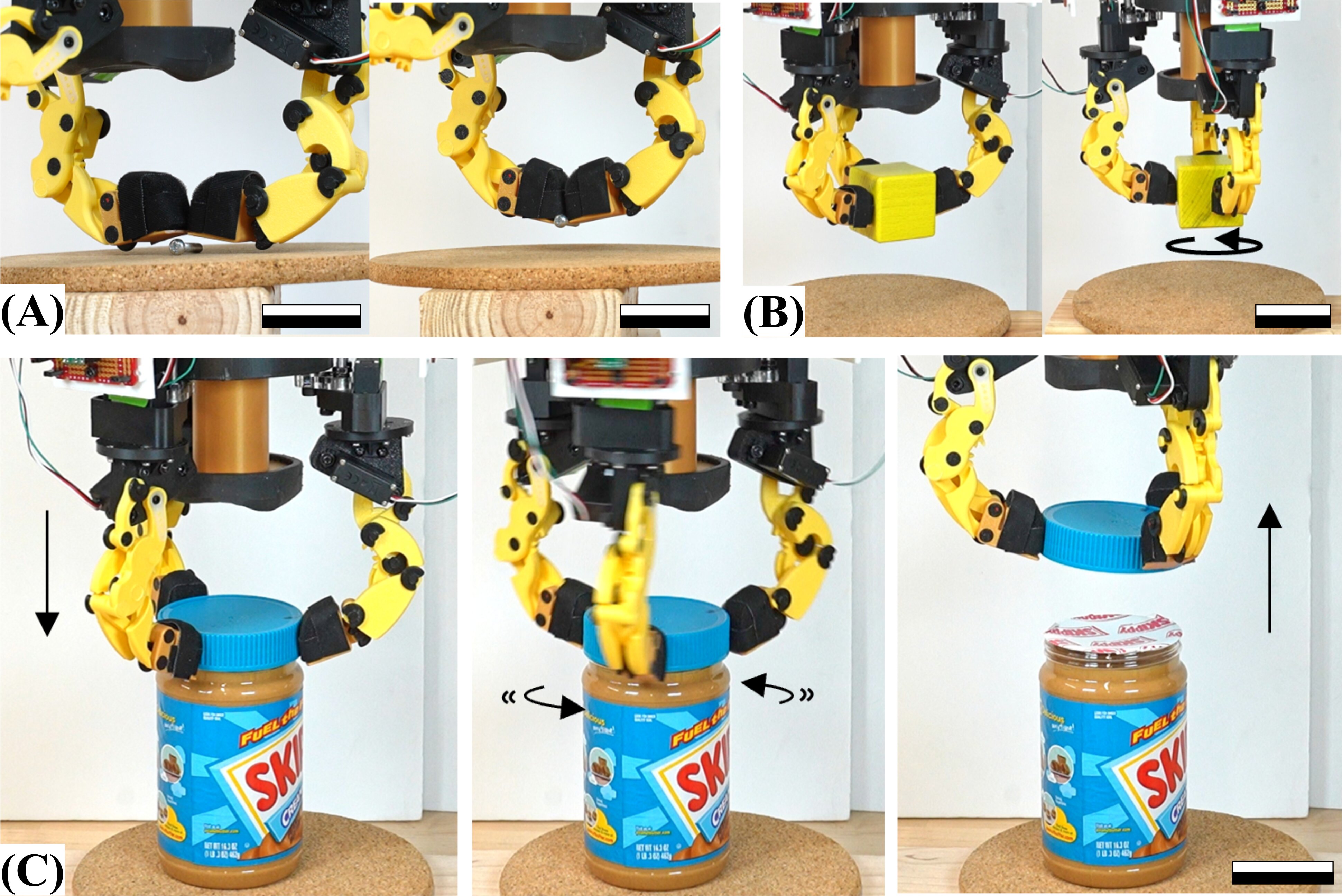}
  \caption{Special tasks using the dynamic orbital DOF~1 (A) Sub-centimeter object pick up (M3x5mm screw), (B) Fast cube orientation, (C) Jar-lid unscrewing.}
  \label{fig:Special habilities}
  \vspace{-5mm}
\end{figure}

Motivated by our failure to pick up the screw, we completed proof-of-concept experiments to demonstrate additional manipulation capabilities that \orbit{}'s rotating knuckle mechanism enables (Fig.~\ref{fig:Special habilities}). Unlike  static hand designs, \orbit{} can use DOF~1 as an active manipulation primitive while the fingers maintain contact with the object. %

First, we performed precision pinching on the M3x\SI{5}{\milli\meter} screw. Unlike our fixed finger root demonstrations, we could successfully grasp the screw by converging two carriages dynamically to the required finger separation before actuating DOF~3. The finger displacement allows the rigid fingernail to make the first contact and slide under the screw head before the sponge pad engages. Second, we performed in-hand cube re-orientation. We grasped the cube in the isometric configuration and sent a synchronized step command to rotate all three carriages by +\qty{30}{\degree}. This allowed us to reorient the cube face by \qty{90}{\degree} in under one second, without the learned multi-finger coordination policies required by other hands  ~\cite{shaw_leap_2023,Siqi_cube_rotation}. Finally, we demonstrate jar-lid unscrewing. With the isometric configuration gripping the peanut butter jar lid, a continuous DOF~1 rotation drove all three carriages counterclockwise, transmitting torque to the lid and unscrewing it in a single motion, demonstrating torque transmission without a dedicated wrist DOF.

Overall, these demonstrations showcase how \orbit{}'s modular features each contributed to grasping performance. The fingernail enabled flat-face interlock and sub-centimeter precision, the sponge pad and friction tape improved surface conformity and grip on curved objects, and the soft palm provided a compliant backstop during power grasps.

\section{Conclusions and future work} 

In this paper, we presented \orbit{}, a modular robotic hand designed as a platform to explore different hand morphologies via a rotating knuckle mechanism that allows dynamic finger reconfiguration. We detailed the design of the rotating knuckle mechanism, presented a 3D-printed two-DOF linkage-based finger design, and characterized the hand across position and velocity for all three DOF per finger. We used \orbit{}'s dynamic reconfigurability to mimic 3 established hand morphologies and directly compare their grasping force and performance against one another. We further highlighted how \orbit{}'s non-anthropomorphic ability to dynamically change finger position during manipulation could achieve further manipulation behavior that is normally difficult to achieve with purely anthropomorphic motions.

For future work, we plan to improve the control of the multi-finger platform with defined grasp commands based on force control and tactile sensing for the fingers. With the goal of exploring hand morphologies more deeply, we aim to simulate the hand platform in software to study the optimization of fundamental variables when designing a robotic hand, such as: the number of fingers, DOFs per finger, DOF configurations, palm designs, and finger designs for particular and generic tasks. With this, our final goal is to physically validate the sim-to-real transfer using \orbit{} as the reconfigurable hardware platform.

\bibliographystyle{IEEEtran}
\bibliography{references_arxiv}

\end{document}